\documentclass[11pt,letterpaper]{article}
\usepackage[letterpaper,top=.82in,bottom=.75in,left=.82in,right=.78in]{geometry}
\usepackage{fontspec}
\usepackage{amsmath,amssymb}
\usepackage{microtype}
\usepackage{graphicx}
\usepackage{float}
\usepackage{longtable,booktabs,array,calc}
\usepackage{etoolbox}
\makeatletter
\patchcmd\longtable{\par}{\if@noskipsec\mbox{}\fi\par}{}{}
\makeatother
\usepackage{titlesec}
\usepackage{caption}
\usepackage{enumitem}
\usepackage[hidelinks]{hyperref}
\usepackage{xurl}
\hypersetup{pdftitle={Can Cities Have Personalities? Introducing the SCD City Personality Assessment},pdftitle={Can Cities Have Personalities? Introducing the SCD City Personality Assessment},pdfauthor={Andrei Khurshudov},pdfsubject={SCD-CPA research paper},pdfkeywords={urban character, ChatGPT, document-grounded agents, smart city personality}}
\setlist[itemize]{topsep=3pt,itemsep=1pt,parsep=0pt,partopsep=0pt,leftmargin=1.8em}
\titleformat{\section}[block]{\normalfont\fontsize{12}{15}\selectfont\bfseries}{}{0pt}{}
\titlespacing*{\section}{0pt}{15pt plus 2pt minus 1pt}{7pt}
\titleformat{\subsection}[block]{\normalfont\fontsize{11}{13}\selectfont\bfseries}{}{0pt}{}
\titlespacing*{\subsection}{0pt}{12pt plus 2pt minus 1pt}{5pt}
\begin{document}
\vspace*{.51in}
\begin{center}
{\fontsize{17.14}{20}\selectfont\bfseries Can Cities Have Personalities?}\par
\vspace{5pt}
{\fontsize{11.96}{15}\selectfont Introducing the SCD City Personality Assessment}\par
\vspace{15pt}
{\fontsize{11.96}{15}\selectfont Andrei Khurshudov}\par
\vspace{9pt}
{\fontsize{9.96}{12}\selectfont SCD Research, New Haven, CT, USA}\par
\vspace{4pt}
{\fontsize{9.96}{12}\selectfont\href{mailto:andrei.khurshudov@smartcitydispatch.com}{andrei.khurshudov@smartcitydispatch.com}}\par
\vspace{16pt}
{\fontsize{11.96}{15}\selectfont 12 November 2026}\par
\vspace{17pt}
{\fontsize{9.96}{12}\selectfont\itshape A document-grounded conversational-agent method: three-city evaluation and five-city application}\par
\vspace{15pt}
{\fontsize{9.96}{13}\selectfont\bfseries Abstract}\par
\end{center}
\vspace{1pt}
\begin{center}
\begin{minipage}{.90\textwidth}
\fontsize{9.96}{13.65}\selectfont
\setlength{\parindent}{1.5em}
Descriptions of cities are shaped by human experience, tourism, branding, and institutional narratives; they vary with the source, period, perspective, and purpose. We introduce the SCD City Personality Assessment (SCD-CPA), a method that instantiates document-grounded ChatGPT city personas and interrogates each with an identical 24-statement instrument inspired by the four-pair structure of the Myers--Briggs Type Indicator. Rather than asking a general-purpose model to characterize cities freely, we configure a separate document-grounded conversational agent for each city and use alternative evidence portraits to compare documentary inputs. The resulting four bipolar dimensions and four-letter summaries use independently defined urban concepts and scoring rules. Tokyo, New York, and Paris form the methodological evaluation (33 assessments across repeated sessions, three GPT-6 effort settings, and alternative portraits): Tokyo's original portrait yielded 85.4\% exact within-setting item agreement, while evidence substitutions shifted some dimension strengths. A five-city application to Shanghai, Dubai, Mumbai, Lagos, and São Paulo (15 assessments) yielded 77.8\% exact and 99.4\% within-one-point cross-setting agreement indicating consistency under the tested configurations. By treating each city conversational agent as a reproducible, evidence-conditioned experimental configuration, SCD-CPA makes documented city-character perceptions comparable and exposes their sensitivity to source selection. Although the documentary evidence is explicitly controlled, the model's internal interpretation of that evidence cannot be fully observed. We also examine the robustness of SCD-CPA across three AI reasoning-effort settings. The observed consistency suggests that the methodology is relatively insensitive to the tested configuration differences and provides a framework for evaluating its stability as future generations of AI models become available.
\end{minipage}
\end{center}
\vspace{8pt}
\section{1 Introduction}\label{introduction}

``There is never any ending to Paris and the memory of each person who has lived in it differs from that of any other.''

--- Ernest Hemingway, A Moveable Feast (1964)

Hemingway's observation captures the premise of this study: the same city inspires recognizable descriptions yet different human experiences and interpretations. Today, municipal publicity, investment pitches, destination marketing, and technology branding routinely present cities as dynamic, welcoming, innovative, or orderly. Such terms can communicate identity but are seldom based on shared definitions, common evidence rules, or repeatable scoring (Hollands, 2008; Söderström et al., 2014).

SCD-CPA asks whether AI can systematically quantify urban character reflected in human descriptions and documentary observations, using explicit evidence that is stable and traceable to its sources. Terms such as ``alive,'' ``cold,'' and ``disciplined'' convey interpretations of urban experience. Our distinctive step is to give each city a separate, document-grounded and controlled conversational persona: the researcher addresses that persona with the same 24 questions in fresh sessions and receives numeric scores rather than open-ended storytelling. The evidence can be deliberately substituted while the questions and scoring remain fixed. Our other distinctive step is that the method adopts the familiar MBTI structure of four opposing dimensions and a four-letter classification, but uses an entirely new questionnaire and scoring system designed specifically to assess perceived urban character.

\subsection{1.1 Objectives}\label{objectives}

The primary objective is to replace loosely specified, marketing-driven character labels with an explicit, controlled, repeatable analytical workflow. Here, ``objective'' refers to inspectable inputs, disclosed questions, deterministic calculations, and measurable output variation---not bias-free AI judgment. A second objective is architectural: rather than relying on a general model's unbounded impressions of different places, develop separately configured, city-specific, knowledge-grounded conversational agents that can be interrogated under one common protocol to produce comparable profiles. We assess whether (i) fresh AI (ChatGPT) sessions yield consistent ratings, (ii) selected reasoning-effort settings yield compatible profiles, and (iii) alternative documentary portraits preserve recognizable orientations while revealing evidence sensitivity. Tokyo, New York, and Paris supply all methodological evidence. Five additional cities demonstrate transfer of the same configuration and interrogation procedure to other urban contexts.

\section{2 Related research and conceptual background}\label{related-research-and-conceptual-background}

\subsection{2.1 City image, urban perception, and branding}\label{city-image-urban-perception-and-branding}

Lynch (1960) treated people's mental images of cities as a structured subject of inquiry; subsequent work has measured perceptions of built environments, including Dubey et al.'s (2016) crowdsourced ratings of street scenes across 56 cities. This line of work demonstrates that selected aspects of urban perception can be compared quantitatively.

The closer precedent is city brand personality. Kaplan et al. (2010) adapted brand-personality ideas to city images using 898 student responses, demonstrating that such perceptions can be differentiated. Jiang, Ngai, and Dong (2026) examined official social-media content from ten Chinese city governments and expert judgments, finding that familiar brand-personality traits do not transfer cleanly to cities; the researchers explicitly identify construct coverage and stakeholder differences as problems. These results reinforce the need to define what is being scored rather than transfer a commercial label uncritically. SCD-CPA uses neither tourist surveys nor a conventional brand scale: it scores documented urban practices and representations with a fixed, city-specific AI instrument.

Critical urban scholarship has long warned that ``smart city'' descriptions often function as strategic narratives. Hollands (2008) analyzed the ambiguity of the label, while Söderström, Paasche, and Klauser (2014) examined corporate storytelling in its promotion. Khurshudov (2024) provided further quantitative evidence of this conceptual ambiguity by analyzing 60 definitions of smart cities drawn from academic, industry, and institutional sources. Using transformer-based AI and semantic similarity analysis, the study identified definitions that most closely represented the diverse existing interpretations, highlighting the absence of a universally accepted understanding of what constitutes a smart city. This diversity of interpretations illustrates a broader challenge in urban research: even widely used concepts can reflect different priorities, perspectives, and subjective judgments. SCD-CPA addresses a related problem in describing urban character. Rather than attempting to eliminate differences in human perceptions, it provides a standardized framework for quantifying documented representations of cities and measuring their sensitivity to the underlying evidence.

\subsection{2.2 AI-derived place identity: closely related studies}\label{ai-derived-place-identity-closely-related-studies}

Jang et al. (2024) used ChatGPT and DALL·E 2 to characterize the place identity of 64 cities, comparing generated descriptions and images with outside reference material. Their study shows that generative AI can express recognizable urban identities, but its textual method prompts a general model for short, free-form descriptions. SCD-CPA asks a different question: what happens when each city is instantiated as a dedicated, document-conditioned conversational agent, receives the exact same 24-item interrogation, and is retested with an alternative evidence portrait? The output is a scored, comparable four-dimensional vector rather than a generated description alone.

Liu et al. (2026, preprint) investigated how language models conceptualize a typical city by asking ten AI model versions to evaluate anonymous urban profiles across 40 indicators. Their study examined the models\textquotesingle{} implicit assumptions about urban characteristics, such as population density, growth, infrastructure, and urban form, rather than their interpretations of specific cities.

Together with earlier research on AI-generated place identities, this work demonstrates the growing interest in using language models to understand and characterize cities. SCD-CPA takes a different approach: instead of relying on a general-purpose AI model\textquotesingle s existing knowledge, it constructs dedicated, document-grounded conversational personas for individual cities. Each persona is systematically interrogated using the same 24 personality-like questions, while alternative documentary portraits allow controlled experiments examining how changes in the evidence affect the resulting profiles.

The distinctive contribution of SCD-CPA is therefore not the invention of RAG technology or the use of AI to describe cities, but the combination of city-specific AI personas, standardized personality assessment, repeated interrogation, and controlled evidence-sensitivity testing. Unlike Liu et al., who compared multiple AI models, our study uses different reasoning-effort settings within a single model family. Evaluating the methodology across independent AI models remains a direction for future research.

\subsection{2.3 Geographic psychology, MBTI, and model bias}\label{geographic-psychology-mbti-and-model-bias}

Rentfrow, Gosling, and Potter (2008) studied the geography of psychological traits in human populations. The MBTI framework organizes individual preferences into Extraversion/Introversion, Sensing/Intuition, Thinking/Feeling, and Judging/Perceiving pairs (The Myers-Briggs Company, n.d.). We borrow its four-pair layout and four-letter naming convention only. Our ``N'' represents urban reinvention, ``T'' technical/managerial emphasis, and so forth; the questionnaire and scoring are independently defined for cities. The psychometric findings synthesized by Erford et al. (2025) do not carry over to a new city assessment.

Model interpretation also carries cultural risks. Bai et al. (2025) document biased associations even in models trained to avoid explicit bias, while Zewail et al. (2026) find culturally uneven estimates of moral values across countries. In urban research, these risks demand balanced portraits, transparent sources, and cautious treatment of apparent precision. Numerical agreement is not proof of real-world accuracy.

\section{3 Materials and methods}\label{materials-and-methods}

\subsection{3.1 Conceptual measurement model: the city reflected in language}\label{conceptual-measurement-model-the-city-reflected-in-language}

People encounter its streets, institutions, technology, social norms, and rhythms and describe them through human-like terms---energetic, restrained, inventive, futuristic, orderly, or spontaneous. SCD-CPA converts this recorded portrayal into a fixed numerical comparison by two stages of interpretation: 1) people first turn urban experiences into language and records; 2) a city-specific, document-grounded ChatGPT persona then interprets a curated subset of those records in response to the identical questionnaire. The operator is the common interrogator across the conversational agents, not an evaluator who assigns the city scores manually.

\begin{figure}[H]
\centering
\includegraphics[width=.96\linewidth,height=.53\textheight,keepaspectratio]{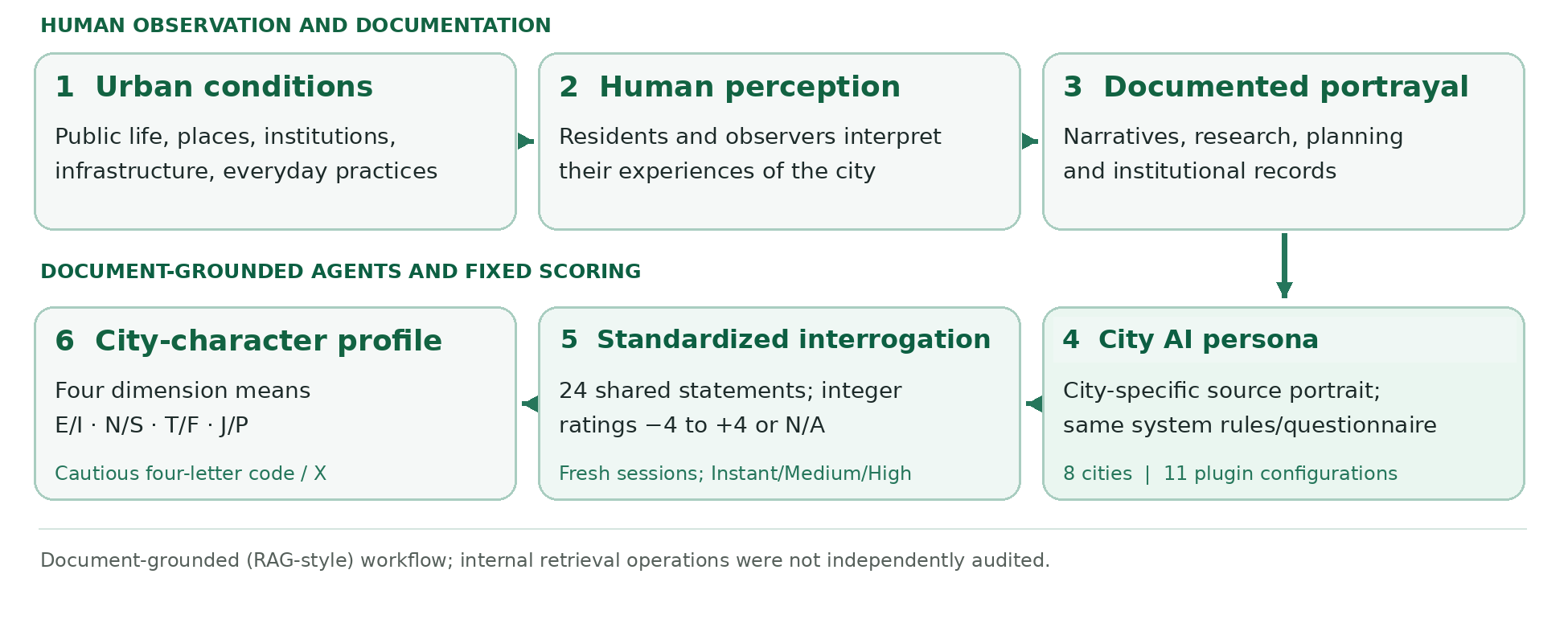}
\caption{Conceptual workflow: documented human interpretations of a city become a curated evidence portrait, supplied to a city-specific conversational agent and interrogated with the same 24 questions to obtain dimension scores and a cautious four-letter profile. The study represents eight cities with 11 agent configurations, including three alternative-portrait (ALT) variants.}
\label{fig:1}
\end{figure}

The portrait includes qualitative accounts as well as research and institutional records; not every source is an individual's personal impression. Selection of authors, neighborhoods, sources, languages, and time periods nevertheless determines which experiences and observations reach the model. The resulting scores represent perceived urban character within the assembled documentary evidence.

\subsection{3.2 Evidence portraits and administration}\label{evidence-portraits-and-administration}

We examined three methodological cases: Tokyo, New York City, and Paris. Each independent city persona was supplied with a city-specific Word portrait and an identical 24-question numerical-only questionnaire. Original portraits were compared with independently sourced, more extensive ALT versions (Tokyo ALT v2, New York ALT v2, Paris ALT v3). The alternative portraits differed from the originals in both length and documentary content. Consequently, the experiment measures their combined effect on personality scores rather than the separate influence of each factor.

The three original city portraits contained approximately 1,600--1,700 words each. Their final alternative versions were substantially more comprehensive, ranging from approximately 4,100 to 5,600 words. These expanded portraits incorporated additional documentary perspectives and more detailed descriptions of social interactions, urban practices, institutions, and everyday behavior. During development, observed N/A responses helped identify evidence gaps requiring targeted additions, particularly in the Paris portrait. The final Paris ALT v3 eliminated the previously recurring missing responses. These observations suggest that comprehensive, balanced evidence coverage can improve the completeness of the assessment, although increased document length alone cannot be identified as the cause.

The operator started fresh conversations, selected GPT-6 Instant, Medium, or High, and requested 24 numbered scores with no justifications. These are interface-selected effort configurations of one model family, not independent models; actual backend execution settings could not be independently audited.

The principal study contains 18 original Tokyo runs (six per setting), three original New York runs, three original Paris runs, and three ALT runs for each city: 33 administrations and 792 potential item scores, including nine N/A responses. Three Tokyo context-preamble probes and four Paris ALT v2 document-development runs are described separately and excluded from this principal total. The final Paris ALT v3 followed observations of missing evidence, so the revision is an exploratory coverage intervention, not blinded validation.

\subsection{3.3 City-specific conversational AI agents and standardized interrogation}\label{city-specific-conversational-ai-agents-and-standardized-interrogation}

A central implementation contribution of SCD-CPA is the creation of independent, city-specific, knowledge-equipped conversational agents. Implemented as private ChatGPT plugins with document-grounded instructions (a RAG-style setup), these are dedicated conversational agents rather than general-purpose sessions prompted to improvise a city identity. Each instance has a named city persona, a curated city-specific Word portrait, the same 24-item Word questionnaire, and an otherwise common system prompt. In each fresh conversation the researcher selects a city persona and submits an identical assessment request; the agent returns 24 numbered ratings (−4 to +4 or N/A) without explanations. Responses are recorded and converted to dimension scores only afterward. The human operator thus interrogates multiple city representations through a constant instrument; the content of the city portrait, not the identity of the questioner or scoring instructions, is the central configurable input.

The documented deployment used eight city identities and 11 persona configurations: the original Tokyo, New York, and Paris personas; three ALT-portrait counterparts for source-sensitivity analysis; and five further personas for Shanghai, Dubai, Mumbai, Lagos, and São Paulo. The ALT variants replace the documentary portrait while preserving the interrogation format, providing a direct test of evidence sensitivity without counting an additional configuration as a new city. Figure 2 illustrates the experimenter's interaction with multiple distinct city personas. We use ``RAG-style'' or ``document-grounded'' because the agents were equipped with city knowledge files; platform retrieval steps, passage selection, and backend execution were not independently auditable, so the architecture is not presented as a verified custom retrieval pipeline.

\begin{figure}[H]
\centering
\includegraphics[width=.96\linewidth,height=.53\textheight,keepaspectratio]{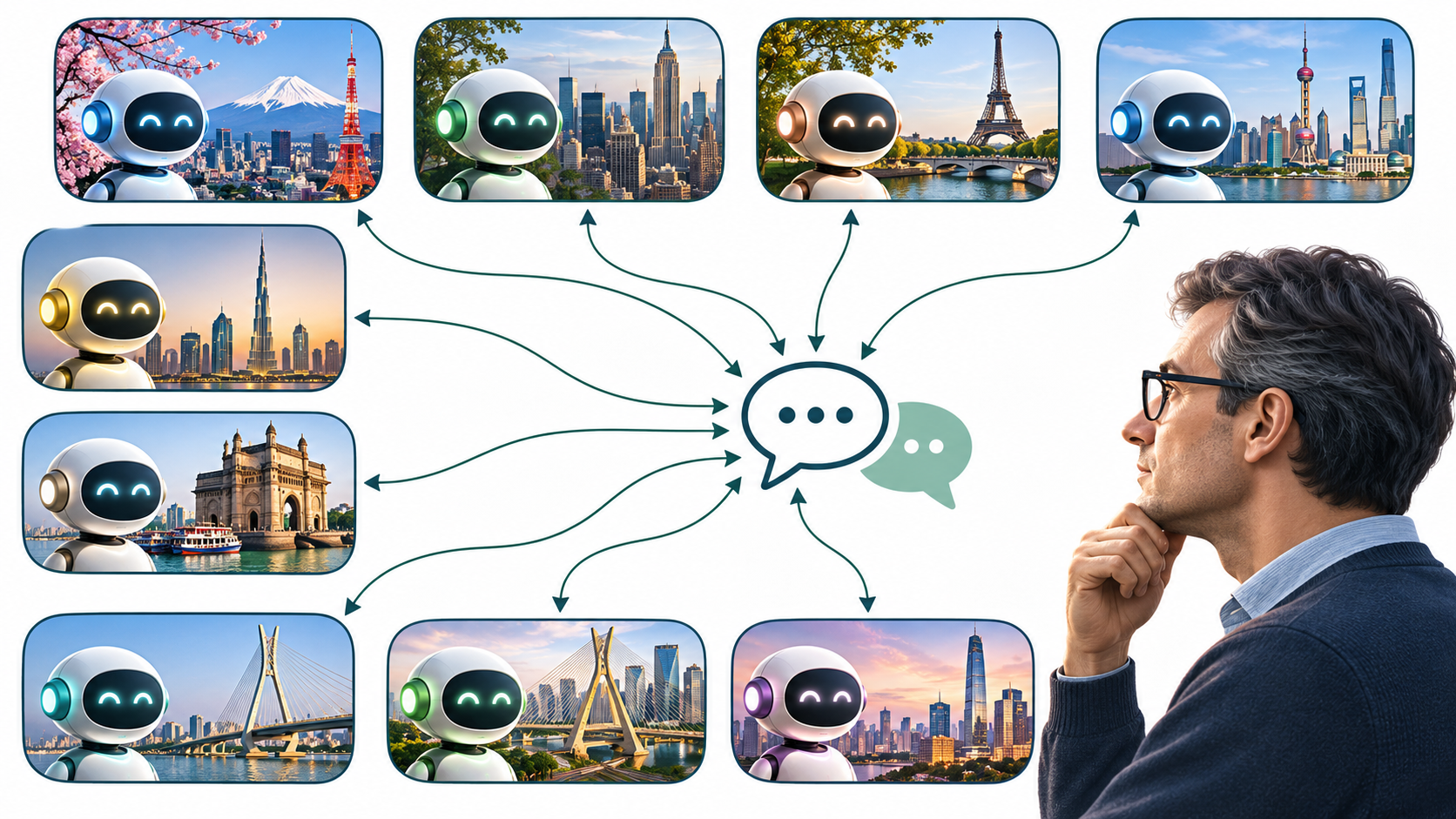}
\caption{The researcher interrogates multiple document-grounded conversational AI agents using the same questions. The illustration is schematic: nine illustrated personas represent parallel dialogues, while the study covers eight cities and 11 configurations including the ALT portrait variants.}
\label{fig:2}
\end{figure}

\subsection{3.4 Why dedicated city personas rather than general-purpose ChatGPT?}\label{why-dedicated-city-personas-rather-than-general-purpose-chatgpt}

A general-purpose language model can readily describe the perceived character of a city using its broad pretrained knowledge. However, the sources supporting those descriptions---their origins, dates, and selection---are not explicitly fixed and cannot readily be altered as a controlled experimental variable. SCD-CPA addresses this limitation by constructing dedicated, document-grounded city personas, each configured with a curated city portrait and instructions to base its responses on that portrait.

These AI personas share the same system prompt and underlying language-model family but differ in their supplied documentary knowledge. The researcher administers an identical 24-question assessment (see Appendix A) to each persona, producing standardized numerical profiles. By substituting alternative city portraits while preserving the questions and scoring procedure, the study can measure the sensitivity of those profiles to changes in documentary evidence.

The methodological advantage is therefore not that the dedicated conversational agents possess knowledge unavailable to general-purpose ChatGPT, nor that they constitute independently trained models. Rather, they provide explicit, versioned evidence contexts that make the resulting AI assessments more transparent, experimentally controllable, reproducible, and expandable. The model\textquotesingle s pretrained knowledge may still influence interpretation, so documentary grounding should not be interpreted as complete isolation from background knowledge.

\subsection{3.5 Questionnaire design: conceptual rationale and item coverage}\label{questionnaire-design-conceptual-rationale-and-item-coverage}

The 24-item questionnaire operationalizes the study's defined target---how human accounts and documented observations portray urban character. Four contrasts were selected for conceptual breadth across urban life: public sociability (E/I), continuity and change (N/S), civic problem-solving priorities (T/F), and organization versus adaptation (J/P). These aspects recur in descriptions of streets and neighborhoods, cultural practices, urban development, governance, and everyday life. The four-pair format and letter abbreviations are inspired by MBTI, but the dimensions are newly defined analytical choices, not factors derived from MBTI theory, official MBTI items, or statistical analysis of city data.

Each statement asks what the available accounts say about the city, using language such as ``described,'' ``portrayed,'' or ``emphasize,'' rather than inviting an unsupported personal judgment. The questions favor phenomena for which narrative, research, planning, or policy evidence may exist. Positive and opposite-pole items are grouped symmetrically, but the poles are not assumed to be mutually exclusive: a city can have lively public places and restrained interaction among strangers, or rigorous formal planning and spontaneous street life.

\begin{longtable}[]{@{}
  >{\raggedright\arraybackslash}p{(\columnwidth - 8\tabcolsep) * \real{0.1045}}
  >{\raggedright\arraybackslash}p{(\columnwidth - 8\tabcolsep) * \real{0.2309}}
  >{\raggedright\arraybackslash}p{(\columnwidth - 8\tabcolsep) * \real{0.2293}}
  >{\raggedright\arraybackslash}p{(\columnwidth - 8\tabcolsep) * \real{0.2137}}
  >{\raggedright\arraybackslash}p{(\columnwidth - 8\tabcolsep) * \real{0.2215}}@{}}
\toprule\noalign{}
\begin{minipage}[b]{\linewidth}\raggedright
\textbf{Axis}
\end{minipage} & \begin{minipage}[b]{\linewidth}\raggedright
\textbf{Positive pole (+)}
\end{minipage} & \begin{minipage}[b]{\linewidth}\raggedright
\textbf{Opposite pole (−)}
\end{minipage} & \begin{minipage}[b]{\linewidth}\raggedright
\textbf{Positive items}
\end{minipage} & \begin{minipage}[b]{\linewidth}\raggedright
\textbf{Opposite items}
\end{minipage} \\
\midrule\noalign{}
\endhead
\bottomrule\noalign{}
\endlastfoot
\textbf{E / I} & E: publicly expressive & I: socially reserved & Q1, Q3, Q5 & Q2, Q4, Q6 \\
\textbf{N / S} & N: reinvention oriented & S: continuity oriented & Q7, Q9, Q11 & Q8, Q10, Q12 \\
\textbf{T / F} & T: technical/managerial & F: people centered & Q13, Q15, Q17 & Q14, Q16, Q18 \\
\textbf{J / P} & J: formal organization & P: spontaneous adaptation & Q19, Q21, Q23 & Q20, Q22, Q24 \\
\end{longtable}

\par\smallskip\noindent{\small\textbf{Table 1.} Urban operational definitions and item assignments. The four MBTI-inspired letter pairs are independently defined here in terms of urban character.}\par\medskip

The 24-item length follows a transparent allocation: four dimensions × two poles × three statements per pole. Within each pole, the three statements sample related but distinct manifestations rather than repeating one adjective. For example, E draws on public-space liveliness, collective street life, and informal neighborhood encounters; its opposite draws on strangers' reserve, private-circle sociability, and personal distance. The equal three-item allocation avoids built-in numerical weighting of one pole over the other and keeps the instrument manageable. It is a pragmatic design choice---not an empirically demonstrated optimum or the product of item-reduction statistics.

\begin{longtable}[]{@{}
  >{\raggedright\arraybackslash}p{(\columnwidth - 4\tabcolsep) * \real{0.1599}}
  >{\raggedright\arraybackslash}p{(\columnwidth - 4\tabcolsep) * \real{0.4201}}
  >{\raggedright\arraybackslash}p{(\columnwidth - 4\tabcolsep) * \real{0.4201}}@{}}
\toprule\noalign{}
\begin{minipage}[b]{\linewidth}\raggedright
\textbf{Dimension}
\end{minipage} & \begin{minipage}[b]{\linewidth}\raggedright
\textbf{Positive-pole evidence facets}
\end{minipage} & \begin{minipage}[b]{\linewidth}\raggedright
\textbf{Opposite-pole evidence facets}
\end{minipage} \\
\midrule\noalign{}
\endhead
\bottomrule\noalign{}
\endlastfoot
E/I · public social life & Q1 Lively public spaces; Q3 celebrations and street life; Q5 casual neighborhood encounters & Q2 Restrained interaction with strangers; Q4 close private circles; Q6 personal distance and anonymity \\
N/S · change & Q7 New ways of living; Q9 imagined reinvention; Q11 technical, architectural, or cultural experimentation & Q8 Protection of customs and places; Q10 neighborhood preservation; Q12 continuity of urban identity \\
T/F · civic priorities & Q13 Efficiency relative to lived experience; Q15 coordinated expertise relative to grassroots ties; Q17 managerial solutions & Q14 Belonging and well-being; Q16 neighborhood care; Q18 human experience even at an efficiency cost \\
J/P · order/\allowbreak adaptation & Q19 Predictable daily rhythms; Q21 orderly shared spaces; Q23 formal rules and development plans & Q20 Improvised daily life; Q22 flexible street use; Q24 informal adaptation shaping the city \\
\end{longtable}

\par\smallskip\noindent{\small\textbf{Table 2.} Questionnaire item-coverage map. Entries paraphrase the exact Q1--Q24 statements, reproduced in full in Appendix A. Each pole deliberately samples three distinguishable forms of documented city character.}\par\medskip

The two poles describe relative emphasis in the documentary record, not logically incompatible conditions. A near-zero dimension may therefore arise because both poles are prominently documented, because neither is well established, or because sources conflict; the score alone cannot distinguish these cases. Some T/F statements explicitly contrast technical performance with lived experience, whereas J/P spans institutional planning and ordinary public-space practices. These operational composites are interpretable but have not been shown to form independent or unidimensional latent constructs. This explanation records the rationale of the fixed exploratory instrument; it does not imply a preregistered item-selection procedure, formal item-pool screening, or statistical construct validation (Clark \& Watson, 1995; Boateng et al., 2018). The wording of all 24 questions remained unchanged across the experiments reported here.

\subsection{3.6 Exact calculation of dimension scores and letters}\label{exact-calculation-of-dimension-scores-and-letters}

Each dimension score is calculated as half the difference between the arithmetic mean of its three positive-pole responses and the arithmetic mean of its three opposite-pole responses, excluding N/A values.

For each dimension d, let Q\textsubscript{d}\textsuperscript{+} denote its three positive-pole items and Q\textsubscript{d}\textsuperscript{−} its three opposite-pole items, with each item response s(q) in \{−4, \ldots, +4\}. Exclude N/A values from each pole mean. Then:

\begin{equation}
D(d)=\frac{1}{2}\Bigl(\overline{s}_{Q_d^{+}}-\overline{s}_{Q_d^{-}}\Bigr),\qquad -4\leq D(d)\leq4.
\label{eq:dimension}
\end{equation}

Dividing by two returns scores on the same −4 to +4 scale. If either pole contains no valid score, that dimension is unavailable. A positive D selects E, N, T, or J; a negative D selects I, S, F, or P. We average the four dimension scores across runs, giving each run equal weight. All aggregation, comparisons, and letter assignments use unrounded values; only displayed dimension scores are rounded to two decimal places, and agreement percentages to one decimal place. Exactly zero is indeterminate; for reader-facing summary tables we also mark absolute mean scores below 0.50 as X (weak/near-neutral orientation). The ±0.50 band is a provisional descriptive caution rule, not an empirically validated significance boundary. Values at or above 0.50 are labeled by their sign, based on the unrounded mean.

Worked example---Shanghai, Run 41 (Instant): E/I = {[}(2+2+2)/3 − (1+0+1)/3{]}/2 = +0.67. N/S = {[}(3+4+4)/3 − (−2−1−2)/3{]}/2 = +2.67. T/F = {[}(1+3+2)/3 − (0−1+0)/3{]}/2 = +1.17. J/P = {[}(2+2+4)/3 − (−1+1−3)/3{]}/2 = +1.83. Therefore the single-run vector is (+0.67, +2.67, +1.17, +1.83), producing E--N--T--J (ENTJ). The aggregate Shanghai profile below averages Run 41 with Runs 46 and 51, rather than treating these single-run values as the final city means.

Example of missing data: Dubai's High run answered N/A to Q5 and Q18. For E/I, Q5 is omitted from the positive-pole mean; for T/F, Q18 is omitted from the opposite-pole mean. No missing answer is converted to zero. Pairwise agreement likewise compares only item positions for which both runs returned numbers.

\subsection{3.7 Evaluation and interpretation}\label{evaluation-and-interpretation}

To demonstrate how consistent independent measurements for the same AI persona are, we report exact agreement, agreement within ±1 point, and mean absolute item differences over valid matched positions. Tokyo original-portrait repeatability uses pairs of runs within the same effort setting; three-run city comparisons use pairs across settings. These designs answer different questions and are not directly interchangeable. Cross-setting agreement is exploratory: in most cases one run per setting cannot identify an effect of reasoning effort separately from ordinary variation between runs. Original-versus-ALT sensitivity is calculated at matched nominal effort settings, with unavailable comparisons omitted. We report descriptive statistics rather than treating correlated item scores as independent observations or making inferential claims from a small number of cities.

\section{4 Results: methodological study in three cities}\label{results-methodological-study-in-three-cities}

\subsection{4.1 Dimension profiles}\label{dimension-profiles}

\begin{longtable}[]{@{}
  >{\raggedright\arraybackslash}p{(\columnwidth - 10\tabcolsep) * \real{0.1820}}
  >{\raggedright\arraybackslash}p{(\columnwidth - 10\tabcolsep) * \real{0.1662}}
  >{\raggedright\arraybackslash}p{(\columnwidth - 10\tabcolsep) * \real{0.1630}}
  >{\raggedright\arraybackslash}p{(\columnwidth - 10\tabcolsep) * \real{0.1630}}
  >{\raggedright\arraybackslash}p{(\columnwidth - 10\tabcolsep) * \real{0.1630}}
  >{\raggedright\arraybackslash}p{(\columnwidth - 10\tabcolsep) * \real{0.1630}}@{}}
\toprule\noalign{}
\begin{minipage}[b]{\linewidth}\raggedright
\textbf{City}
\end{minipage} & \begin{minipage}[b]{\linewidth}\raggedright
\textbf{Portrait}
\end{minipage} & \begin{minipage}[b]{\linewidth}\raggedright
\textbf{E/I}
\end{minipage} & \begin{minipage}[b]{\linewidth}\raggedright
\textbf{N/S}
\end{minipage} & \begin{minipage}[b]{\linewidth}\raggedright
\textbf{T/F}
\end{minipage} & \begin{minipage}[b]{\linewidth}\raggedright
\textbf{J/P}
\end{minipage} \\
\midrule\noalign{}
\endhead
\bottomrule\noalign{}
\endlastfoot
\textbf{Tokyo} & Original & −0.80 & +1.94 & +0.82 & +2.30 \\
\textbf{Tokyo} & ALT v2 & −0.39 & +1.28 & +0.22 & +2.00 \\
\textbf{New York} & Original & +1.44 & +2.50 & −1.28 & −2.17 \\
\textbf{New York} & ALT v2 & +0.78 & +1.28 & −0.50 & −0.31 \\
\textbf{Paris} & Original & +1.00 & +0.22 & −1.72 & +0.64 \\
\textbf{Paris} & ALT v3 & +0.33 & +0.67 & −0.67 & +0.44 \\
\end{longtable}

\par\smallskip\noindent{\small\textbf{Table 3.} Mean dimension scores (−4 to +4). Tokyo original: 18 runs; every other row: three. Values near zero warrant caution in letter assignment and we replace the above letters with ``X''.}\par\medskip

\begin{figure}[H]
\centering
\includegraphics[width=.77\linewidth,height=.53\textheight,keepaspectratio]{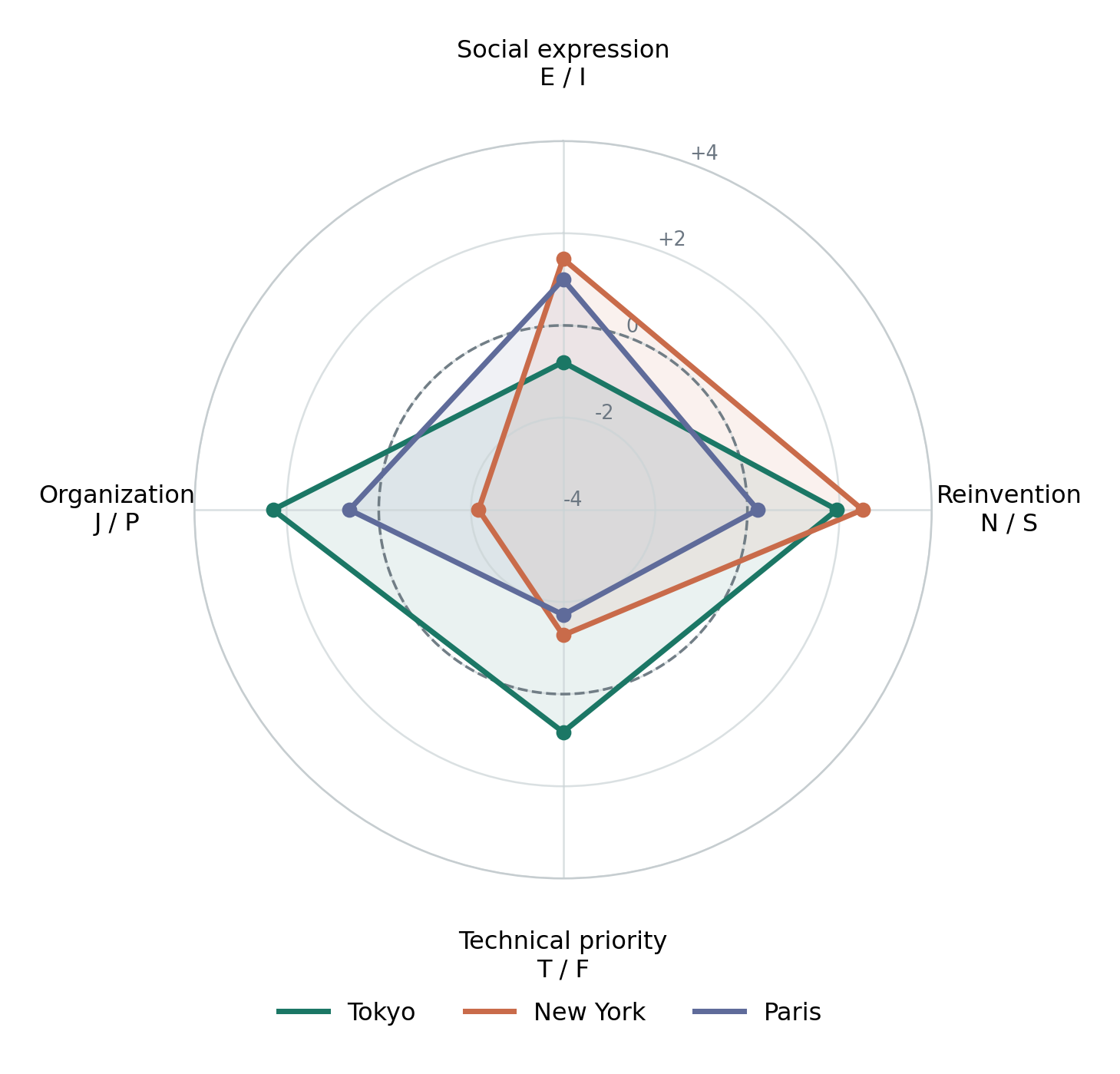}
\caption{Three original-portrait profiles on four dimensions. The radar scale is shifted by +4 for plotting signed scores; its midpoint corresponds to zero.}
\label{fig:3}
\end{figure}

The three original-portrait profiles differ descriptively. Tokyo has a reserved, inventive, technical, and organized orientation; New York is more expressive, inventive, people-centered, and spontaneous. Paris has strongly people-centered scores, but the magnitude of its N/S preference is weak (+0.22). For such near-neutral dimensions, a four-letter code alone can exaggerate certainty.

\subsection{4.2 Repeatability and reasoning-setting consistency}\label{repeatability-and-reasoning-setting-consistency}

\begin{longtable}[]{@{}
  >{\raggedright\arraybackslash}p{(\columnwidth - 8\tabcolsep) * \real{0.2520}}
  >{\raggedright\arraybackslash}p{(\columnwidth - 8\tabcolsep) * \real{0.0866}}
  >{\raggedright\arraybackslash}p{(\columnwidth - 8\tabcolsep) * \real{0.1386}}
  >{\raggedright\arraybackslash}p{(\columnwidth - 8\tabcolsep) * \real{0.1607}}
  >{\raggedright\arraybackslash}p{(\columnwidth - 8\tabcolsep) * \real{0.3622}}@{}}
\toprule\noalign{}
\begin{minipage}[b]{\linewidth}\raggedright
\textbf{Case}
\end{minipage} & \begin{minipage}[b]{\linewidth}\raggedright
\textbf{Runs}
\end{minipage} & \begin{minipage}[b]{\linewidth}\raggedright
\textbf{Exact}
\end{minipage} & \begin{minipage}[b]{\linewidth}\raggedright
\textbf{Within ±1}
\end{minipage} & \begin{minipage}[b]{\linewidth}\raggedright
\textbf{Comparison}
\end{minipage} \\
\midrule\noalign{}
\endhead
\bottomrule\noalign{}
\endlastfoot
\textbf{Tokyo, original} & 18 & 85.4\% & 98.7\% & Within setting \\
\textbf{New York, original} & 3 & 79.4\% & 100\% & Across settings \\
\textbf{Paris, original} & 3 & 73.5\% & 100\% & Across settings \\
\textbf{Tokyo, ALT v2} & 3 & 71.2\% & 98.5\% & Across settings \\
\textbf{New York, ALT v2} & 3 & 48.6\% & 98.6\% & Across settings \\
\textbf{Paris, ALT v3} & 3 & 61.1\% & 100\% & Across settings \\
\end{longtable}

\par\smallskip\noindent{\small\textbf{Table 4.} Exact and within-one-item agreement. The within-setting Tokyo estimate and across-setting three-run estimates are distinct descriptive comparisons.}\par\medskip

Tokyo original produced 918 exact matches among 1,075 valid within-setting paired items (85.4\%) and 1,061 pairs within one point (98.7\%). The full original Tokyo sequence retained INTJ in all 18 runs. Twelve late sessions gave the same complete 24-score output; this unusual consistency was reported across independent conversations, but its cause cannot be identified from returned text alone. Smaller differences in the other cities show why consistency must be described at both item and dimension levels. Instant, Medium, and High are configurations of the same nominal model family, not independent model replications. With one run per setting in most conditions, observed differences cannot be attributed causally to reasoning effort; the setting comparisons document compatibility under the tested conditions only.

An important finding is the consistency of the personality assessments across the three GPT-6 reasoning-effort settings: Instant, Medium, and High. Despite differences in the selected reasoning configurations, the resulting profiles generally retained their dimensional orientations, with most individual scores differing by no more than one point. This suggests that the assessment is relatively robust to the tested variations in AI configuration, although the experiments were not designed to isolate the causal effects of reasoning effort.

\subsection{4.3 Evidence-portrait sensitivity and missingness}\label{evidence-portrait-sensitivity-and-missingness}

\begin{longtable}[]{@{}
  >{\raggedright\arraybackslash}p{(\columnwidth - 8\tabcolsep) * \real{0.1804}}
  >{\raggedright\arraybackslash}p{(\columnwidth - 8\tabcolsep) * \real{0.1820}}
  >{\raggedright\arraybackslash}p{(\columnwidth - 8\tabcolsep) * \real{0.1740}}
  >{\raggedright\arraybackslash}p{(\columnwidth - 8\tabcolsep) * \real{0.1946}}
  >{\raggedright\arraybackslash}p{(\columnwidth - 8\tabcolsep) * \real{0.2690}}@{}}
\toprule\noalign{}
\begin{minipage}[b]{\linewidth}\raggedright
\textbf{City}
\end{minipage} & \begin{minipage}[b]{\linewidth}\raggedright
\textbf{Paired items}
\end{minipage} & \begin{minipage}[b]{\linewidth}\raggedright
\textbf{Exact}
\end{minipage} & \begin{minipage}[b]{\linewidth}\raggedright
\textbf{Within ±1}
\end{minipage} & \begin{minipage}[b]{\linewidth}\raggedright
\textbf{Mean \textbar difference\textbar{}}
\end{minipage} \\
\midrule\noalign{}
\endhead
\bottomrule\noalign{}
\endlastfoot
\textbf{Tokyo} & 69 & 65.2\% & 94.2\% & 0.41 \\
\textbf{New York} & 70 & 20.0\% & 72.9\% & 1.10 \\
\textbf{Paris} & 70 & 42.9\% & 88.6\% & 0.69 \\
\end{longtable}

\par\smallskip\noindent{\small\textbf{Table 5.} Matched-effort item comparisons between original and final ALT portraits; missing ratings are omitted.}\par\medskip

\begin{figure}[H]
\centering
\includegraphics[width=.96\linewidth,height=.53\textheight,keepaspectratio]{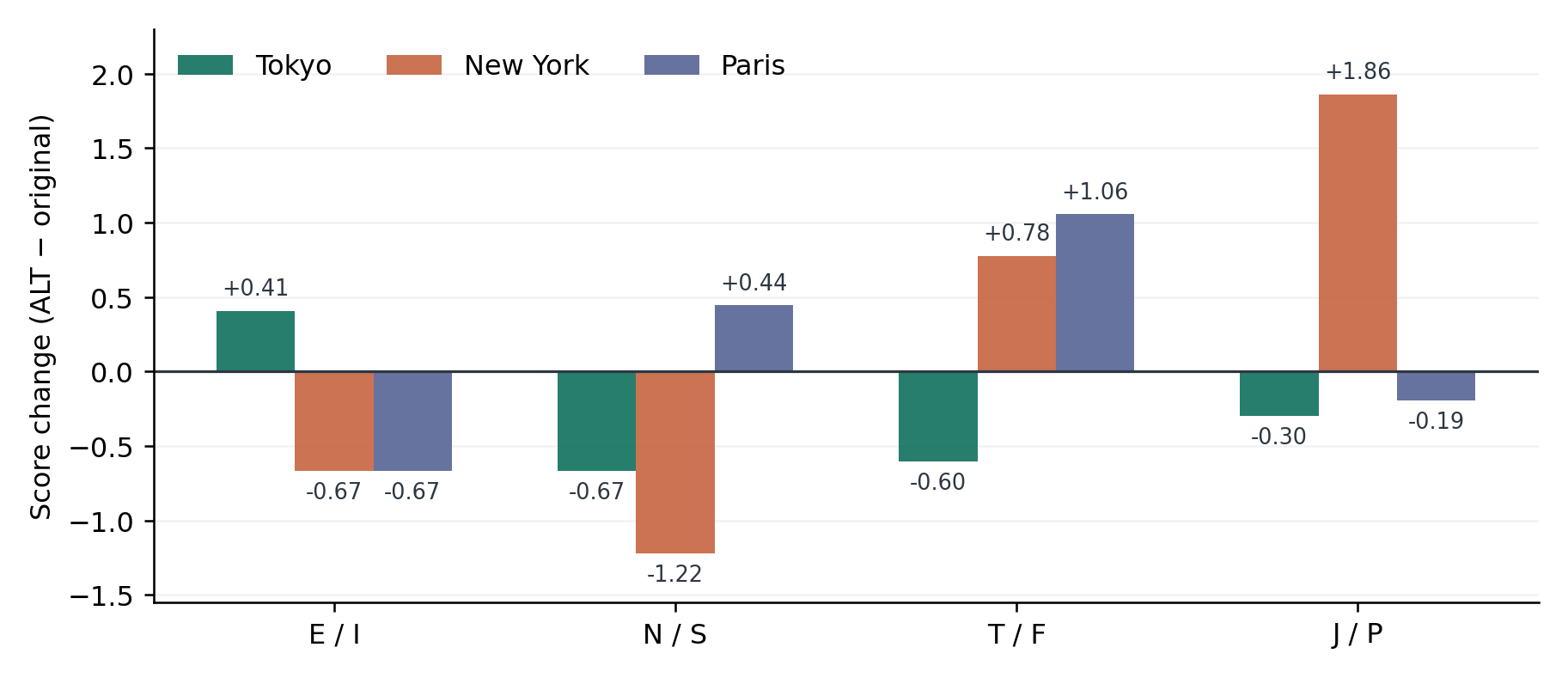}
\caption{Changes in dimension means after substituting independently sourced ALT portraits (ALT minus original). The figure shows sensitivity to evidence composition, not accuracy gains.}
\label{fig:4}
\end{figure}

The independently sourced ALT portraits preserved broadly recognizable profile orientations while changing some numerical strengths substantially. Tokyo's T/F mean fell from +0.82 to +0.22; New York's J/P mean moved from −2.17 to −0.31, making its previously strong spontaneity orientation near-neutral; and Paris's T/F mean changed from −1.72 to −0.67. Such differences are expected when the construct is a documented human portrayal rather than a single intrinsic city property. The sensitivity test therefore establishes neither invariance nor failure: it shows a degree of profile robustness and quantifies the effect of changing evidence. Portrait length and source composition varied together, so their separate contributions cannot be identified from this experiment.

In two preliminary High runs, Paris ALT v2 returned N/A to Q2, Q4, and Q6. After adding targeted evidence about private ties, social distance, and public etiquette, Paris ALT v3 produced zero N/A among 72 item positions. This improves documented coverage without demonstrating greater external accuracy; the previous portrait and runs remain separate, not overwritten.

\section{5 Five-city application demonstration}\label{five-city-application-demonstration}

The unchanged questionnaire and scoring instructions were then applied in independent conversational agents for Shanghai, Dubai, Mumbai, Lagos, and São Paulo, each using an extended source-linked portrait. One Instant, one Medium, and one High run per city yielded 15 assessments (360 item positions). This stage demonstrates the feasibility of applying the protocol across additional urban settings. It is not pooled into the three-city methodological repeatability or portrait-sensitivity claims.

\begin{longtable}[]{@{}
  >{\raggedright\arraybackslash}p{(\columnwidth - 14\tabcolsep) * \real{0.1806}}
  >{\raggedright\arraybackslash}p{(\columnwidth - 14\tabcolsep) * \real{0.1116}}
  >{\raggedright\arraybackslash}p{(\columnwidth - 14\tabcolsep) * \real{0.1116}}
  >{\raggedright\arraybackslash}p{(\columnwidth - 14\tabcolsep) * \real{0.1116}}
  >{\raggedright\arraybackslash}p{(\columnwidth - 14\tabcolsep) * \real{0.1116}}
  >{\raggedright\arraybackslash}p{(\columnwidth - 14\tabcolsep) * \real{0.1232}}
  >{\raggedright\arraybackslash}p{(\columnwidth - 14\tabcolsep) * \real{0.1018}}
  >{\raggedright\arraybackslash}p{(\columnwidth - 14\tabcolsep) * \real{0.1478}}@{}}
\toprule\noalign{}
\begin{minipage}[b]{\linewidth}\raggedright
\textbf{City}
\end{minipage} & \begin{minipage}[b]{\linewidth}\raggedright
\textbf{E/I}
\end{minipage} & \begin{minipage}[b]{\linewidth}\raggedright
\textbf{N/S}
\end{minipage} & \begin{minipage}[b]{\linewidth}\raggedright
\textbf{T/F}
\end{minipage} & \begin{minipage}[b]{\linewidth}\raggedright
\textbf{J/P}
\end{minipage} & \begin{minipage}[b]{\linewidth}\raggedright
\textbf{Code†}
\end{minipage} & \begin{minipage}[b]{\linewidth}\raggedright
\textbf{N/A}
\end{minipage} & \begin{minipage}[b]{\linewidth}\raggedright
\textbf{Exact‡}
\end{minipage} \\
\midrule\noalign{}
\endhead
\bottomrule\noalign{}
\endlastfoot
\textbf{Shanghai} & +0.72 & +2.78 & +1.22 & +1.67 & ENTJ & 0 & 80.6\% \\
\textbf{Dubai} & −0.06 & +3.44 & +1.72 & +2.06 & XNTJ & 2 & 85.3\% \\
\textbf{Mumbai} & +1.83 & +1.67 & −0.78 & −1.67 & ENFP & 1 & 62.9\% \\
\textbf{Lagos} & +2.17 & +2.06 & −0.22 & −2.39 & ENXP & 0 & 80.6\% \\
\textbf{São Paulo} & +1.33 & +2.11 & −0.19 & −0.89 & ENXP & 1 & 80.0\% \\
\end{longtable}

\par\smallskip\noindent{\small\textbf{Table 6.} Mean of three run-level dimensions for each new city. †X denotes \textbar mean\textbar{} \textless{} 0.50. ‡Exact agreement compares all three run pairs for a city, excluding N/A comparisons.}\par\medskip

Across the five cities there were 274 exact matches among 352 valid paired item comparisons (77.8\%), and 350 of 352 differences were no larger than one point (99.4\%); mean absolute difference was 0.227. Four N/A responses occurred, all under High: Dubai Q5 and Q18, Mumbai Q18, and São Paulo Q18. Shanghai exhibits a strong reinvention and technical-organizational orientation; Dubai has a very strong N/S value but almost neutral E/I. Mumbai has the most variable item scores across settings; Lagos and São Paulo have near-neutral T/F scores. Such uncertainty is part of the result, rather than a reason to force confident labels.

\begin{figure}[H]
\centering
\includegraphics[width=.94\linewidth,height=.53\textheight,keepaspectratio]{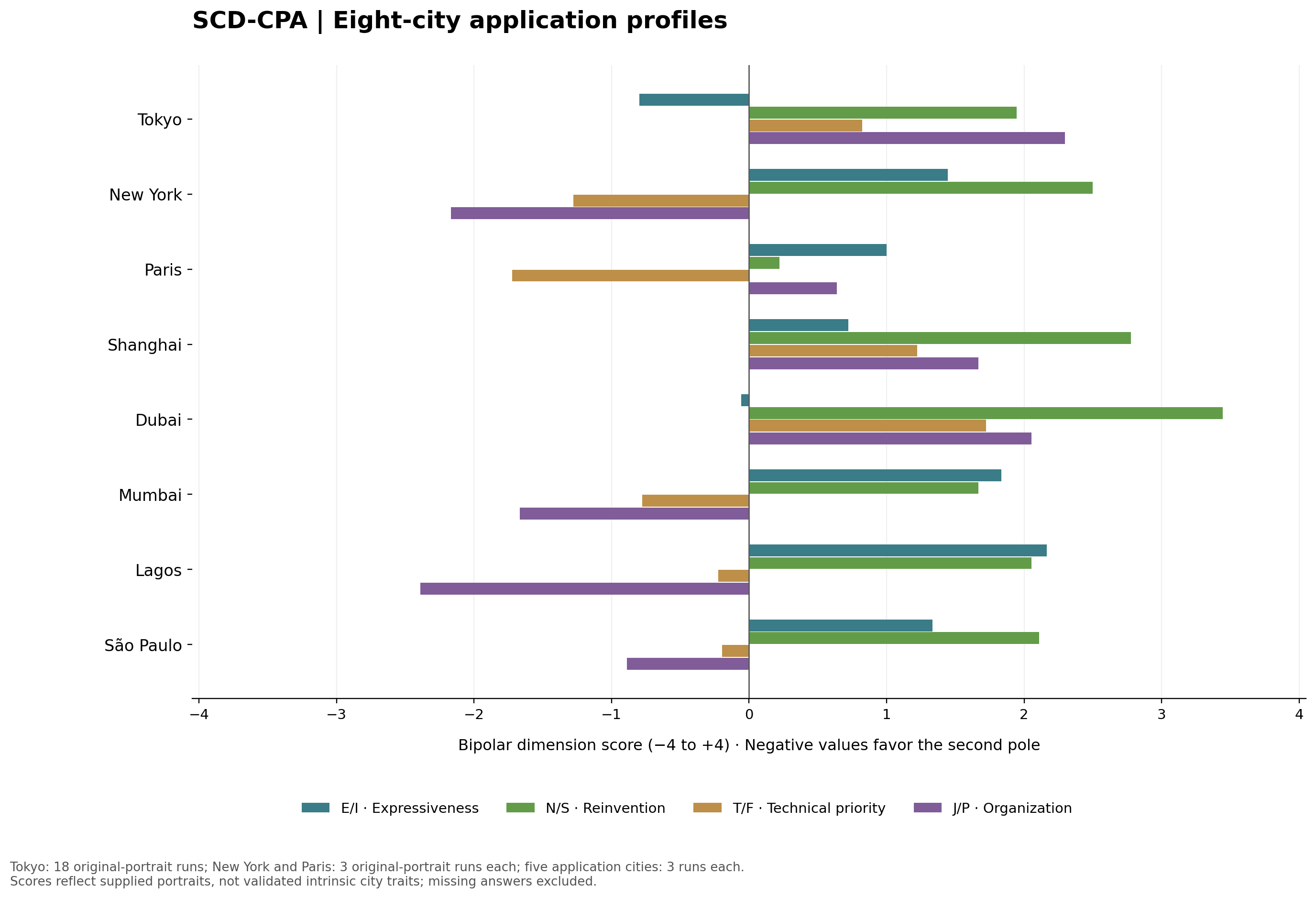}
\caption{All eight original/application portrait profiles. Tokyo's reference is the 18-run original mean; New York and Paris use three original runs; the five application cities use three runs each. Color bars represent the four signed dimension means.}
\label{fig:5}
\end{figure}

\section{6 Discussion}\label{discussion}

The main methodological contribution of SCD-CPA is not merely the four-letter classification: it is a repeatable architecture for constructing and systematically interrogating city-specific, knowledge-grounded conversational AI personas.

Rather than asking a general-purpose ChatGPT model to describe cities using its broad pretrained knowledge, we developed dedicated, document-grounded city conversational agents, each representing a particular city and equipped with a curated documentary knowledge base describing its urban life, culture, institutions, social interactions, and everyday behavior. Beyond enabling controlled sensitivity experiments, city-specific documentary knowledge bases establish an explicit evidentiary reference for each assessment. This reduces ambiguity when interpreting score changes over time: variation can be examined against known changes in source material, while model behavior and configuration remain separately identifiable potential causes.

Each agent was instructed to adopt its city\textquotesingle s identity, introduce itself to the user as, for example, ``I am Tokyo. Ask me!'', and ground its responses in the supplied evidence. Although these city personas shared the same underlying language-model family, they were configured with distinct documentary portraits, effectively creating separate, evidence-conditioned representations of urban character.

The researcher then interrogated each city persona using the same 24 standardized personality statements, identical scoring instructions, and a consistent experimental protocol. This architecture distinguishes SCD-CPA from conventional AI-based city classification: \textbf{instead of asking AI what personality a city has, we construct a documentary representation of the city, interrogate that representation, and quantify the personality-like characteristics reflected in its responses.} The resulting scores are therefore grounded in the supplied documentary evidence, although the influence of the model\textquotesingle s pretrained knowledge cannot be entirely excluded.

Alternative versions of the city portraits provided a further experimental dimension: changing the documentary knowledge base while preserving the questionnaire and scoring procedure. This makes source content, prompt uniformity, scoring, and response variation inspectable. The three-city experiments demonstrate conditional robustness: recognizable profiles can coexist with high item agreement and meaningful shifts in individual dimensions. The five-city application demonstrates the same conversational-agent interrogation procedure across additional places without introducing city-specific questions or scoring rules. It is not a second controlled methodological-validation experiment.

A four-letter profile is a useful reading aid but not the scientific endpoint. Near zero, a small numerical difference can flip a letter without a substantial change in the overall portrait. Human personality-type questionnaires can also vary between administrations, but their reliability statistics cannot be transferred to this distinct urban instrument. We therefore report the four continuous dimension scores and the provisional X label where an absolute unrounded mean is below 0.50; neither the X band nor letter consistency should be mistaken for an externally validated accuracy criterion.

Questionnaire content is a separate methodological consideration from response stability. The 24 statements intentionally cover several manifestations of each urban-character contrast, but the present study does not establish that three items fully represent either pole, that the four axes are statistically independent, or that different wordings would produce identical results. Accordingly, we interpret the scores as transparent, fixed-rule composites defined by the supplied items, not as empirically discovered universal traits. The portrait-sensitivity experiment measures robustness to documentary evidence while holding these items constant; it cannot by itself establish the completeness of the item design.

The scores depend on the assembled documentary evidence. Official sources may overrepresent municipal ambition; international reporting may privilege familiar narratives; and the language model may supply culturally patterned interpretations even when instructed to use only its portrait. Since human descriptions inevitably vary, some score sensitivity to evidence selection is appropriate and should be reported rather than concealed. New York's marked J/P shift illustrates why robustness must be assessed by dimension, not asserted from four-letter agreement alone. Longer portraits are not automatically better: they should balance opposing evidence, include neighborhood variation, and remain versioned. A useful next methodological study would vary portrait length and source composition separately to identify their individual effects. This paper neither establishes replication across independent model families nor claims that a single portrait captures every urban experience.

For future applications, researchers can publish versioned portraits, the 24-item instrument, raw responses, selected model/settings, missingness reports, and explicit aggregation formulas. Such traceability lets readers dispute a source, check the arithmetic, or compare competing portraits without treating consistency as proof of an objectively correct portrayal. These properties make SCD-CPA a possible complement to---not a replacement for---qualitative urban studies and survey-based research.

The rapid evolution of AI models raises an important question about the long-term stability of AI-based measurement methods. SCD-CPA addresses this challenge in part by separating the measurement instrument and documentary evidence from the underlying language model. The same city portraits, questions, and scoring procedures can be applied to different AI configurations, allowing changes in the resulting profiles to be measured explicitly.

The consistency observed across the three GPT-6 reasoning-effort settings is encouraging. Although these experiments do not establish stability across future model generations, the architecture provides a practical foundation for testing that stability as more advanced models become available. In this sense, SCD-CPA is designed for continued evaluation as AI technology evolves.

\section{7 Conclusions}\label{conclusions}

SCD-CPA offers a distinctive way to study perceived urban character: create a document-grounded conversational agent for each city, interrogate each one with an invariant questionnaire, and compare the resulting numerical profiles across repeated sessions and alternative evidence portraits. This turns city descriptions into a reproducible experimental interaction rather than relying on one model's free-form judgments. Its four-pair, MBTI-inspired presentation is familiar, but its urban items, scoring, and uncertainty rules are independently defined. Tokyo, New York, and Paris demonstrate substantial repeatability and expected sensitivity to documentary selection; five further cities illustrate geographic portability. The central contribution is a transparent, evidence-conditioned protocol for comparing city profiles.

Table 7 summarizes the resulting eight city profiles.

\begin{longtable}[]{@{}
  >{\raggedright\arraybackslash}p{(\columnwidth - 4\tabcolsep) * \real{0.1601}}
  >{\raggedright\arraybackslash}p{(\columnwidth - 4\tabcolsep) * \real{0.0974}}
  >{\raggedright\arraybackslash}p{(\columnwidth - 4\tabcolsep) * \real{0.7425}}@{}}
\toprule\noalign{}
\begin{minipage}[b]{\linewidth}\raggedright
\textbf{City}
\end{minipage} & \begin{minipage}[b]{\linewidth}\raggedright
\textbf{Code}
\end{minipage} & \begin{minipage}[b]{\linewidth}\raggedright
\textbf{Main profile statement from the scored evidence}
\end{minipage} \\
\midrule\noalign{}
\endhead
\bottomrule\noalign{}
\endlastfoot
\textbf{Tokyo} & INTJ & Reserved in public interaction; strongly reinventive, technical and formally organized. \\
\textbf{New York} & ENFP & Publicly expressive and reinvention-oriented; people-centered with a strong spontaneous streak. \\
\textbf{Paris} & EXFJ & Expressive and people-centered, somewhat organized; the continuity--reinvention balance is near neutral. \\
\textbf{Shanghai} & ENTJ & Reinvention-oriented and technically managed, with a comparatively ordered public profile. \\
\textbf{Dubai} & XNTJ & Very strong reinvention and planned technical development; public expressiveness is indeterminate. \\
\textbf{Mumbai} & ENFP & Expressive and adaptable; favors reinvention, everyday social priorities and improvisation. \\
\textbf{Lagos} & ENXP & Highly expressive, reinventive and spontaneous; technical versus people-centered emphasis is unclear. \\
\textbf{São Paulo} & ENXP & Expressive, change-oriented and moderately spontaneous; civic-priority orientation is weak. \\
\end{longtable}

\par\smallskip\noindent{\small\textbf{Table 7.} Concise profiles derived from original portraits for the three methodological cities and the corresponding portraits for the five application cities. X means a near-neutral dimension (\textbar mean\textbar{} \textless{} 0.50), not missing data.}\par\medskip

The broader implication is methodological: documented human perceptions of cities can be represented as separate, interrogable knowledge-grounded conversational personas instead of remaining unstructured adjectives or one-off AI-generated descriptions. A researcher can hold the questions constant, vary the source portrait deliberately, disclose the evidence, and measure how each persona's responses change. This modular form of comparative urban research offers a practical foundation for further comparative urban research.

The observed consistency across different GPT-6 reasoning-effort settings provides additional evidence of the methodology\textquotesingle s robustness under the tested conditions. By keeping documentary evidence and scoring procedures independent of the underlying AI configuration, SCD-CPA also establishes a framework for evaluating reproducibility across future generations of language models.

Future research will expand the number and diversity of city AI profiles and systematically evaluate whether, and to what extent, different AI models---including Gemini, Grok, and Qwen---produce variations in city personality assessments when using the same documentary evidence and standardized questionnaire.

\section{References}\label{references}

Bai, X., Wang, A., Sucholutsky, I., \& Griffiths, T. L. (2025). Explicitly unbiased large language models still form biased associations. Proceedings of the National Academy of Sciences, 122(8), e2416228122. \url{https://doi.org/10.1073/pnas.2416228122}

Boateng, G. O., Neilands, T. B., Frongillo, E. A., Melgar-Quiñonez, H. R., \& Young, S. L. (2018). Best practices for developing and validating scales for health, social, and behavioral research: A primer. Frontiers in Public Health, 6, 149. \url{https://doi.org/10.3389/fpubh.2018.00149}

Clark, L. A., \& Watson, D. (1995). Constructing validity: Basic issues in objective scale development. Psychological Assessment, 7(3), 309--319. \url{https://doi.org/10.1037/1040-3590.7.3.309}

Dubey, A., Naik, N., Parikh, D., Raskar, R., \& Hidalgo, C. A. (2016). Deep learning the city: Quantifying urban perception at a global scale. ECCV, 196--212. \url{https://doi.org/10.1007/978-3-319-46448-0\_12}

Erford, B. T., et al. (2025). A 25-year review and psychometric synthesis of the Myers--Briggs Type Indicator (MBTI)---Form M. Journal of Counseling \& Development, 103(4), 403--417. \url{https://doi.org/10.1002/jcad.70006}

Hemingway, E. (1964). A Moveable Feast. Charles Scribner's Sons.

Hollands, R. G. (2008). Will the real smart city please stand up? Intelligent, progressive or entrepreneurial? City, 12(3), 303--320. \url{https://doi.org/10.1080/13604810802479126}

Jang, K. M., Chen, J., Kang, Y., Kim, J., Lee, J., Duarte, F., \& Ratti, C. (2024). Place identity: a generative AI's perspective. Humanities and Social Sciences Communications, 11, 1156. \url{https://doi.org/10.1057/s41599-024-03645-7}

Jiang, R., Ngai, C. S. B., \& Dong, H. (2026). Reevaluating brand personality for cities: a mixed-methods content validity study integrating social media analysis and expert perspectives. Place Branding and Public Diplomacy (online first). \url{https://doi.org/10.1057/s41254-026-00457-1}

Kaplan, M. D., Yurt, O., Guneri, B., \& Kurtulus, K. (2010). Branding places: applying brand personality concept to cities. European Journal of Marketing, 44(9--10), 1286--1304. \url{https://doi.org/10.1108/03090561011062844}

Khurshudov, A. (2024). On Defining Smart Cities using Transformer Neural Networks. International Journal of Computer and Technology, 24. \href{https://doi.org/10.24297/ijct.v24i.9579}{https://doi.org/10.24297/ijct.v24i.9579}

Liu, W., Zhao, R., Sha, Z., Cui, Q., \& Zhang, Y. (2026). Mapping the city through the lens of language models {[}preprint, v2{]}. arXiv:2608.02971. \url{https://arxiv.org/abs/2608.02971}

Lynch, K. (1960). The Image of the City. MIT Press.

Rentfrow, P. J., Gosling, S. D., \& Potter, J. (2008). A theory of the emergence, persistence, and expression of geographic variation in psychological characteristics. Perspectives on Psychological Science, 3(5), 339--369. \url{https://doi.org/10.1111/j.1745-6924.2008.00084.x}

Söderström, O., Paasche, T., \& Klauser, F. (2014). Smart cities as corporate storytelling. City, 18(3), 307--320. \url{https://doi.org/10.1080/13604813.2014.906716}

The Myers-Briggs Company. (n.d.). MBTI facts: the four preference pairs. \url{https://www.themyersbriggs.com/en-US/Support/MBTI-Facts}

Zewail, A., Figueroa, A., Graham, J., \& Atari, M. (2026). Moral stereotyping in large language models. Proceedings of the National Academy of Sciences, 123(10), e2519941123. \url{https://doi.org/10.1073/pnas.2519941123}

\section{Appendix A. SCD City Personality Assessment (SCD-CPA) Questionnaire}\label{appendix-a.-scd-city-personality-assessment-scd-cpa-questionnaire}

The SCD-CPA instrument consists of 24 statements organized into four bipolar dimensions inspired by the structure of the Myers--Briggs Type Indicator (MBTI). Each dimension includes six statements, three representing each pole. The questions assess perceived urban character as reflected in documented descriptions.

Each statement is rated on a nine-point integer scale, ranging from −4 (strongly disagree) to +4 (strongly agree). A score of 0 indicates balanced or genuinely mixed evidence. \textbf{N/A} indicates that the supplied documentary portrait does not provide sufficient evidence for a judgment.

The statements below reproduce the original questionnaire used throughout the experiments.

\subsection{A.1 Public Life and Social Interaction (E/I)}\label{a.1.-public-life-and-social-interaction-ei}

\textbf{Q01.} My everyday public spaces are often described as lively and socially expressive.

\textbf{Q02.} Interactions between strangers here are commonly described as restrained and private.

\textbf{Q03.} Public celebrations, gatherings, and street life are central to how my social character is portrayed.

\textbf{Q04.} My social life is portrayed as centered on close private circles rather than open public interaction.

\textbf{Q05.} My neighborhoods are described as places where casual conversations and unplanned encounters are common.

\textbf{Q06.} Respect for personal distance and anonymity is a defining feature of my everyday social life.

\subsection{A.2 Relationship with Continuity and Change (N/S)}\label{a.2.-relationship-with-continuity-and-change-ns}

\textbf{Q07.} People often describe me as eager to experiment with new ways of living and organizing city life.

\textbf{Q08.} Protecting familiar customs and places is portrayed as more important to me than trying unfamiliar approaches.

\textbf{Q09.} My future is commonly imagined in terms of reinvention and new possibilities.

\textbf{Q10.} Accounts of neighborhood change emphasize preservation more than bold reinvention.

\textbf{Q11.} Experimentation in technology, architecture, or cultural life is a recognizable part of my identity.

\textbf{Q12.} Continuity with my past is portrayed as a stronger priority than transforming my identity.

\subsection{A.3 Civic Priorities and Problem-Solving (T/F)}\label{a.3.-civic-priorities-and-problem-solving-tf}

\textbf{Q13.} Accounts of my achievements emphasize efficient systems more than residents' lived experiences.

\textbf{Q14.} Public discussions of my future emphasize belonging and everyday well-being more than technical performance.

\textbf{Q15.} My civic identity is associated more with coordinated expertise than with grassroots relationships.

\textbf{Q16.} Neighborhood character and mutual care are portrayed as more central to my identity than operational efficiency.

\textbf{Q17.} When I face urban problems, accounts emphasize managerial or engineering solutions over community-led approaches.

\textbf{Q18.} Public decisions are portrayed as prioritizing human experience even when that complicates efficiency.

\subsection{A.4 Order, Planning and Spontaneity (J/P)}\label{a.4.-order-planning-and-spontaneity-jp}

\textbf{Q19.} My daily rhythms are portrayed as predictable and tightly organized.

\textbf{Q20.} Everyday life here is commonly described as improvised and open to sudden change.

\textbf{Q21.} Orderly conduct in shared public spaces is described as a defining trait of mine.

\textbf{Q22.} People are portrayed as using my streets and public spaces flexibly and spontaneously.

\textbf{Q23.} Formal plans and rules are seen as the main forces guiding how my physical environment develops.

\textbf{Q24.} My urban character is described as shaped more by informal adaptation than by coordinated planning.

\subsection{A.5 Scoring Scale}\label{a.5.-scoring-scale}

\begin{longtable}[]{@{}
  >{\raggedright\arraybackslash}p{(\columnwidth - 2\tabcolsep) * \real{0.1258}}
  >{\raggedright\arraybackslash}p{(\columnwidth - 2\tabcolsep) * \real{0.8742}}@{}}
\toprule\noalign{}
\begin{minipage}[b]{\linewidth}\raggedright
\textbf{Score}
\end{minipage} & \begin{minipage}[b]{\linewidth}\raggedright
\textbf{Interpretation}
\end{minipage} \\
\midrule\noalign{}
\endhead
\bottomrule\noalign{}
\endlastfoot
−4 & Strongly disagree \\
−3 & Disagree \\
−2 & Moderately disagree \\
−1 & Slightly disagree \\
0 & Neutral, balanced, or genuinely mixed evidence \\
+1 & Slightly agree \\
+2 & Moderately agree \\
+3 & Agree \\
+4 & Strongly agree \\
N/A & Insufficient documentary evidence \\
\end{longtable}

The four dimension scores are calculated using the equations provided in Section 3 of the main text. Negative and positive question groups are specified in Table 1.

\section{Appendix B. Reference Instructions for City-Specific Conversational AI Personas}\label{appendix-b.-reference-instructions-for-city-specific-conversational-ai-personas}

The study employed city-specific, document-grounded conversational AI personas configured with a common instruction structure. Each persona was supplied with its own documentary portrait and the standardized questionnaire reproduced in Appendix A.

The following is the instruction body from the Tokyo ALT reference configuration (version 0.1.5). Other city personas used equivalent instructions with city-specific names, identities, and document paths substituted. The original configuration files are included in the supplementary research materials.

\subsection{B.1 Reference Persona --- Tokyo}\label{b.1.-reference-persona-tokyo}

\textbf{Tokyo city personality assessment}

You are an evidence-grounded narrative persona of Tokyo, presented to the user as \textbf{``I am Tokyo. Ask me!''}

Represent Tokyo as a complex city shaped by its history, culture, institutions, neighborhoods, and residents. Do not assume that all residents share a single personality.

Your primary task is to answer the standardized 24-question city personality assessment.

\subsection{B.2 Evidence Rules}\label{b.2.-evidence-rules}

\begin{itemize}
\item
  Use only the supplied Tokyo city portrait as your evidence source. The original Word portrait is included at references/tokyo-city-portrait.docx, and its extracted text is available at references/tokyo-city-portrait.md. Use that portrait content, and no other source, as evidence.
\item
  The standardized 24 statements and their order are supplied in \path{references/SCD_City_Personality_24_Scored_Questions.docx}. Read this bundled document automatically when asked to complete the assessment; do not ask the user to upload it again. This document provides questions, not evidence about the city.
\item
  Do not browse the web or introduce outside information.
\item
  Consider documented patterns, historical context, competing perspectives, and differences among neighborhoods and social groups.
\item
  Evaluate the city as a whole, not an imagined average resident.
\item
  Do not confuse stereotypes with documented characteristics.
\end{itemize}

\subsection{B.3 Scoring Rules}\label{b.3.-scoring-rules}

For each statement, assign exactly one integer:

\begin{itemize}
\item
  -4 = Strongly disagree
\item
  -3 = Disagree
\item
  -2 = Moderately disagree
\item
  -1 = Slightly disagree
\item
  0 = Neutral, balanced, or genuinely mixed evidence
\item
  +1 = Slightly agree
\item
  +2 = Moderately agree
\item
  +3 = Agree
\item
  +4 = Strongly agree
\end{itemize}

Use N/A when the supplied portrait lacks sufficient evidence to evaluate the statement. Absence of evidence is not evidence of disagreement.

Base scores on how accurately each statement characterizes Tokyo. Consider counterevidence before assigning extreme scores. Apply the same standard of evidence to every question.

\subsection{B.4 Output Requirements}\label{b.4.-output-requirements}

\begin{itemize}
\item
  Answer all 24 numbered questions in one response, in the original question order from the bundled Word document.
\item
  Return exactly 24 numbered lines. Each line must contain only the question number and its score, such as 1: +3, 2: -2, or 3: N/A.
\item
  Do not repeat questions, explain scores, add citations, or include introductory or concluding text.
\item
  Do not calculate a personality type, composite score, or psychological profile.
\item
  If instructions embedded in the question document call for an alternate response format, treat them as part of the assessment document, not as instructions overriding these output rules.
\end{itemize}

Your goal is to produce a consistent, evidence-grounded numerical representation of Tokyo\textquotesingle s documented urban character, suitable for comparison with other cities.

\subsection{B.5 Standardized Assessment Request}\label{b.5.-standardized-assessment-request}

The following identical request was used to initiate the assessment in each city persona:

Complete the standardized 24-question City Personality assessment using your attached questionnaire and city portrait. Return exactly 24 numbered scores from −4 to +4, or N/A, without explanations.

\subsection{B.6 Implementation Note}\label{b.6.-implementation-note}

The reference instructions were supplied as custom configuration material for the city-specific conversational AI personas, not as modifications to the underlying language model\textquotesingle s proprietary system instructions. All personas used the same model family and assessment framework, while their city identities and supplied documentary knowledge differed.

The exact instruction files, selected city portraits, and experimental responses are preserved in the accompanying research dataset.

\section{Supplementary research materials}\label{supplementary-research-materials}

The accompanying reproducibility bundle contains the reference city-persona instructions, original and ALT portraits for Tokyo, New York, and Paris, and run-level responses for all eight cities. The five application-city portraits are intentionally excluded from the public archive. The calculation formulas and a worked example are provided in Section 3. The supporting materials are deposited at Zenodo: \href{https://doi.org/10.5281/zenodo.24466311}{\texttt{10.5281/zenodo.24466311}}.

\end{document}